\documentclass[a4paper, 10pt, conference]{ieeeconf}      

\IEEEoverridecommandlockouts                              

\usepackage{graphics} 
\usepackage{epsfig} 
\usepackage{mathptmx} 
\usepackage{hyperref}
\usepackage{float}
\usepackage{times} 
\usepackage{amsmath} 
\usepackage{amssymb}  
\usepackage{caption}
\usepackage{subcaption}
\title{\LARGE \bf A comparative evaluation of learned feature descriptors on hybrid monocular visual SLAM methods}

\author{Hudson M. S. Bruno $^{1}$ and Esther L. Colombini $^{1}$
\thanks{$^{1}$All authors are with Institute of Computing, State University of Campinas - Unicamp, Av. Albert Einstein, 1251 - Cidade Universitária, Campinas-SP, Brazil.
{\tt\small hudson.bruno@ic.unicamp.br}}%
}

\begin{document}

\maketitle
\thispagestyle{empty}
\pagestyle{empty}

\begin{abstract}
Classical Visual Simultaneous Localization and Mapping (VSLAM) algorithms can be easily induced to fail when either the robot's motion or the environment is too challenging. The use of Deep Neural Networks to enhance VSLAM algorithms has recently achieved promising results, which we call hybrid methods. In this paper, we compare the performance of hybrid monocular VSLAM methods with different learned feature descriptors. To this end, we propose a set of experiments to evaluate the robustness of the algorithms under different environments, camera motion, and camera sensor noise. Experiments conducted on KITTI and Euroc MAV datasets confirm that learned feature descriptors can create more robust VSLAM systems.
\end{abstract}

\section{INTRODUCTION}

\label{sec:introduction}

Visual Odometry (VO) is a process to estimate the pose (position and orientation) of a camera from a sequence of images. If the algorithm estimates the pose and simultaneously reconstructs a three dimensional model of the environment, the process is called Visual Simultaneous Localization and Mapping (VSLAM).

Traditional VSLAM approaches usually follow a fundamental pipeline composed of initialization, tracking, mapping, global optimization, and relocalization. However, these approaches tend to fail in challenging environments such as when the camera moves at high speed or when the image is distorted (due to rolling shutter effect, unfavorable exposure conditions, etc.). Moreover, if the camera is monocular, these systems have scale uncertainty.

Recent developments on deep learning show that pose estimation can be treated as a learning problem \cite{survey-dynamic-envs}. Deep learning-based methods in VO and VSLAM can bring robustness to the situations mentioned above. There are two main approaches to deep learning-based VO/VSLAM systems: end-to-end and hybrid strategies. In end-to-end methods, the deep neural network (DNN) replaces the entire VO pipeline \cite{deep-vo, undeep-vo, attention-based}. However, end-to-end approaches are still incapable of achieving the same performance and accuracy as traditional state-of-the-art approaches. To solve this problem, some authors proposed to split the VSLAM pipeline and use DNNs to execute specific tasks, which we call hybrid strategies \cite{cnn-relocalisation, pose-graph-optimization, monocular-depth}.

Recent work have also proposed detecting and describing image local features with DNNs ~\cite{lift, lf-net, superpoint}.  These learned features can be used as input to VSLAM systems \cite{self-improving-vo, gcnv2}. However, to the best of our knowledge, a comparative evaluation of learned feature descriptors in feature-based hybrid VSLAM pipelines still was not presented in the literature.

\begin{figure}[t]
      \centering
      \includegraphics[scale=0.18]{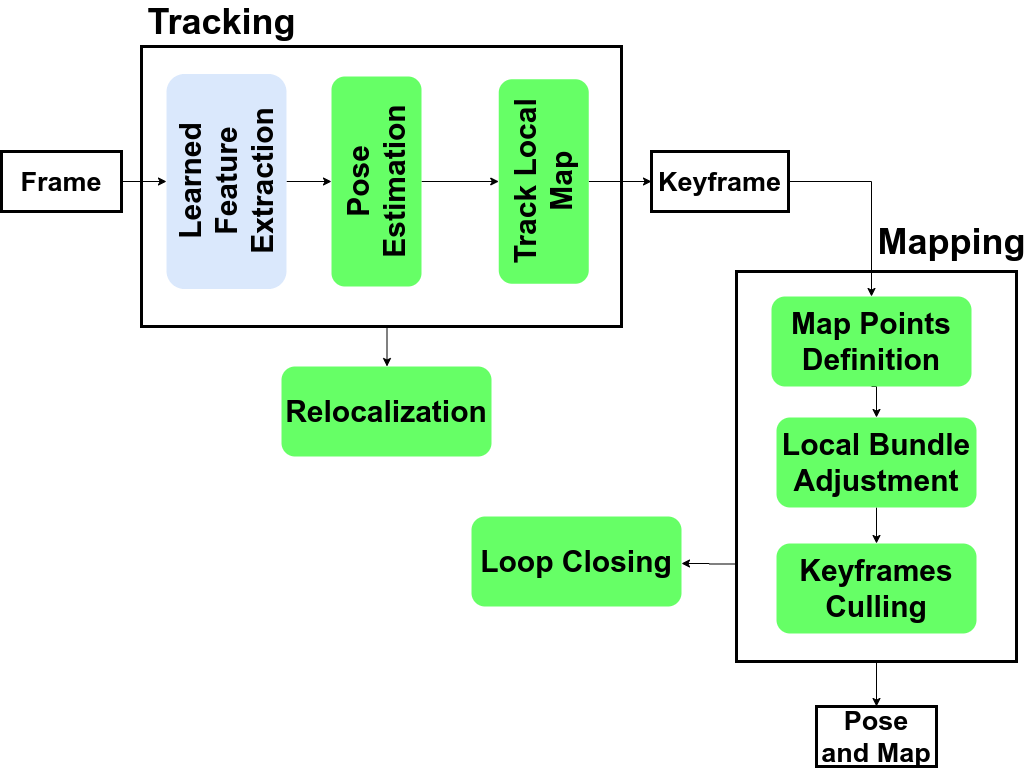}
      \caption{The hybrid VSLAM pipeline used to evaluate the learned feature descriptors effect. The green boxes are the modules from a traditional VSLAM pipeline.}
      \label{fig:evaluation-pipeline}
\end{figure}

Therefore, in this paper, we evaluate the effect of using different learned feature descriptors in the pose estimation of a hybrid monocular VSLAM pipeline, shown in Figure \ref{fig:evaluation-pipeline}. We evaluate two learned descriptors on the same VSLAM traditional back-end, similar to the proposed in the well-known ORB-SLAM \cite{orb-slam} algorithm. We adopted the following learned feature extractors and descriptors: Learned Invariant Feature Transform (LIFT) \cite{lift} and Local Feature Network (LF-Net) \cite{lf-net}. 

This paper presents two main contributions: (1) a hybrid monocular VSLAM pipeline that can adopt any learned feature descriptor in the front-end and an ORB-SLAM's based back-end, (2) a novel approach to evaluate the robustness of hybrid VSLAM algorithms based on KITTI \cite{kitti-dataset} and Euroc MAV \cite{euroc-mav} datasets with the application of camera distortions in some scenarios.

\section{RELATED WORK}
\label{sec:related-work}

\textbf{Geometry-based approaches.}
Geometry-based methods rely on geometric constraints extracted from images to estimate motion. Geometry-based VO/VSLAM algorithms are either direct or feature-based (indirect). Direct algorithms directly use the whole image (its pixel intensities) for tracking and mapping. Some of the state-of-the-art direct VO/VSLAM algorithms are Large-scale Direct SLAM (LSD-SLAM) \cite{lsd-slam} and Direct Sparse Odometry (DSO) \cite{dso}. LSD-SLAM represents maps with semi-dense inverse depth maps for selected keyframes, containing depth values for all pixels with a sufficient intensity gradient. They provide a probabilistic solution to handle the noisy depth prediction during tracking.
Moreover, pose-graph optimization is employed to obtain a geometrically consistent map. On the other hand, DSO combines photometric errors with geometric errors and performs a joint optimization of all parameters. The algorithm divides the input image into a grid and then selects high-intensity points as reconstruction candidates to its sparse map. By using this strategy, they can get points that are distributed within the entire image.

Instead of using the entire image, algorithms based on tracking and mapping feature points are called feature-based approaches. One of the main feature-based algorithms is ORB-SLAM \cite{orb-slam}. It is a monocular VSLAM system that works with three threads: Tracking, Local Mapping, and Loop Closure. It relies on ORB features and uses a place recognition system based on Bag-of-Words (BoW). The mapping step adopts graph representations, which allow the system to perform local and global pose-graph optimization. Later, the authors of ORB-SLAM proposed two extensions of ORB-SLAM: ORB-SLAM2 \cite{orb-slam2} to stereo and RGB-D cameras and ORB-SLAM3 \cite{orb-slam3} to visual-inertial SLAM and multi-map SLAM.

\textbf{Deep learning-based approaches.} There are two main ways of using deep learning to perform VO/VSLAM. In end-to-end approaches, a DNN replaces all modules of a traditional VO pipeline. Wang et. al. proposed one of the most notable deep-learning end-to-end VO approaches, called DeepVO \cite{deep-vo}. In DeepVO, a Recurrent Neural Network (RNN) estimates the camera pose from features learned by a Convolutional Neural Network (CNN). Two stacked Long-Short Term Memory (LSTM) layers are applied to estimate temporal changes from the features predicted by the CNN. Another end-to-end approach, based on unsupervised learning called UnDeepVO, is presented by Li et. al. in \cite{undeep-vo}. The network relies on stereo image pairs to recover the scale during training while using consecutive monocular images for testing.

Hybrid approaches were also proposed to enhance only some modules of the traditional VO/VSLAM pipeline. Li et. al. proposed a monocular system called Neural Bundler  \cite{pose-graph-optimization}. It is an unsupervised DNN that estimates motion. It constructs a conventional pose graph, enabling an efficient loop closing procedure based on the pose graph optimization. Another hybrid VSLAM system was proposed by Tang et. al. (GCNv2) \cite{gcnv2}. It is a deep learning-based network for the generation of keypoints and descriptors. The network is designed with a binary descriptor vector, working as a replacement to ORB in ORB-SLAM2. Furthermore, a self-supervised approach called SuperPointVO is proposed by DeTone et. al. in \cite{self-improving-vo}. In this work, they combine a DNN based in SuperPoint \cite{superpoint} feature extractor as a VSLAM front-end with a traditional back-end, using the stability of keypoints in the images to aid in learning.

\section{METHODOLOGY}
In this section, we give more details about the learned feature descriptors used in this paper. We also present a brief explanation of the traditional VSLAM pipeline adopted in the back-end. Lastly, we discuss the KITTI and Euroc datasets characteristics and the distortions we made in the images to test the robustness of the algorithms. 

\subsection{Feature descriptors}
\textbf{LIFT} \cite{lift} is a DNN that implements local feature detection, orientation estimation, and description. Its architecture comprises three main modules: Detector, Orientation Estimator, and Descriptor, and is trained in a supervised end-to-end approach. The network input are patches of images. The detector gets these patches and outputs keypoints locations while the orientation estimator predicts an orientation to the patch. Lastly, the descriptor computes the local descriptor from the rotated patch. The ground-truth is obtained with a Structure from Motion (SfM) algorithm that is used to reconstruct from image sets using SIFT features \cite{sift}. The version of LIFT used in this paper was provided by the authors in their github\footnote{\url{https://github.com/cvlab-epfl/tf-lift}}. In this version, the network takes approximately 36 seconds to generate the descriptors for a image from the KITTI dataset and 35 seconds for a image from Euroc dataset in a GTX 1050 Ti.

The LIFT parameters used to execute the VSLAM algorithm were: 
\begin{itemize}
    \item Scale factor between levels in the scale pyramid: $2$;
    \item Number of levels in the scale pyramid: $3$;
    \item Matching thresholds: $TH_{LOW} = 1$ and $TH_{HIGH} = 2$ for Euroc sequences and $TH_{LOW} = 2$ and $TH_{HIGH} = 3$ for KITTI sequences.
\end{itemize}

\textbf{LF-Net} \cite{lf-net} is a DNN that also implements local feature detection, orientation estimation, and description. However, this architecture is trained in a self-supervised end-to-end approach by exploiting depth, and relative camera poses from sets of images. The DNN is composed of two main components: the detector and the descriptor. The detector is a CNN that returns keypoint locations, scales, and orientations. The descriptor gets patches cropped around the keypoints and outputs local descriptors. In this paper, we used the indoor and outdoor models of LF-Net, also provided by the authors in their github\footnote{\url{https://github.com/vcg-uvic/lf-net-release}}. In this version, the network takes approximately 0.35 seconds to generate the descriptors for a image from the KITTI dataset and 0.31 seconds for a image from Euroc dataset in a GTX 1050 Ti.

The LF-Net parameters used in the VSLAM algorithm were: 

\begin{itemize}
    \item Scale factor between levels in the scale pyramid: $2$;
    \item Number of levels in the scale pyramid: $\sqrt{2}$;
    \item Matching thresholds: $TH_{LOW} = 1$ and $TH_{HIGH} = 2$ for Euroc sequences and $TH_{LOW} = 2$ and $TH_{HIGH} = 3$ for KITTI sequences.
\end{itemize}

\subsection{VSLAM pipeline}

To evaluate the descriptors' impact in VO/VSLAM, we used a common VSLAM pipeline for both descriptors. This pipeline is based on ORB-SLAM \cite{orb-slam} and is composed of Tracking, Mapping, Loop Closure, and Relocalization. However, as opposed to ORB-SLAM, the only task we run in parallel is loop closure detection.

\textbf{Tracking.} In this step, after extracting the features, a feature matching operation is performed. This operation removes matching outliers, based on distance thresholds. Then, the camera pose is predicted with a constant velocity model. This pose estimation is then optimized by searching for more map point correspondences in the current frame by projecting the local map 3D points into the image. Besides, this step decides whether the frame is a keyframe; if so, the mapping step is performed.

\textbf{Mapping.} This step inserts the keyframe into a co-visibility graph, where keyframes are nodes, and the edges are computed based on the shared map points with other keyframes. Then, new map points are created by triangulating the features from connected keyframes in the graph. Moreover, the co-visibility graph is optimized with a local bundle adjustment algorithm. This algorithm is applied to all keyframes connected to the current keyframe in the co-visibility graph and all map points seen by those keyframes. Lately, we discard redundant keyframes to improve the co-visibility graph's size. 

\textbf{Loop closure and relocalization.} To perform both steps we created visual vocabularies with the DBoW2 library \cite{dbow2}, as proposed in ORB-SLAM \cite{orb-slam}. We created a visual vocabulary for each learned feature descriptor from approximately 12,000 images collected from outdoor and indoor sequences from the TUM-mono VO dataset \cite{tum-mono-vo}. Therefore, each vocabulary has approximately $1,000,000$ visual words, distributed in $6$ levels and $10$ clusters per level, as suggested in \cite{vocabulary-based-slam}. 

The loop closure task gets the last keyframe processed by the local mapping and detects if it closes a loop. For each keyframe, a Bag of Words (BoW) is computed based on the visual vocabulary. This BoW is used to compute a similarity score between the current keyframe and its neighbors in the co-visibility graph. The loop closure is found if there are at least three candidates detected in the same co-visibility graph. Then, a rigid-body transformation is computed from the candidate keyframe to the loop keyframe. This transformation is used to correct the loop.

If the tracking is lost, the relocalization module is activated. It queries the BoW of the current frame into a database composed of a hierarchical BoW to find keyframe candidates for global relocalization.

\subsection{Datasets}
\label{sec:datasets}
Two datasets were chosen to evaluate our algorithms: KITTI \cite{kitti-dataset} and Euroc MAV \cite{euroc-mav}. The KITTI dataset is a collection of images recorded from a moving car, and Euroc MAV is a set of images collected by Micro Aerial Vehicles indoors. Therefore, the datasets have different environments (e.g., outdoor/indoor, size, illumination, dynamic/static, etc.), and camera motion (e.g., acceleration, velocities, DoF, etc.), which allow us to evaluate the robustness of the algorithms to different situations.

Moreover, we created different image distortion in a sequence of KITTI. This allowed us to evaluate the algorithms' robustness to camera sensor noise, simulating camera ill exposure conditions.
We simulated camera overexposure and underexposure with the application of gamma power transformation, as proposed in \cite{emulate-exposure}. 

\begin{figure}[tb]
\centering
\subfloat[][$\gamma < 1$.]{
\includegraphics[scale=0.12]{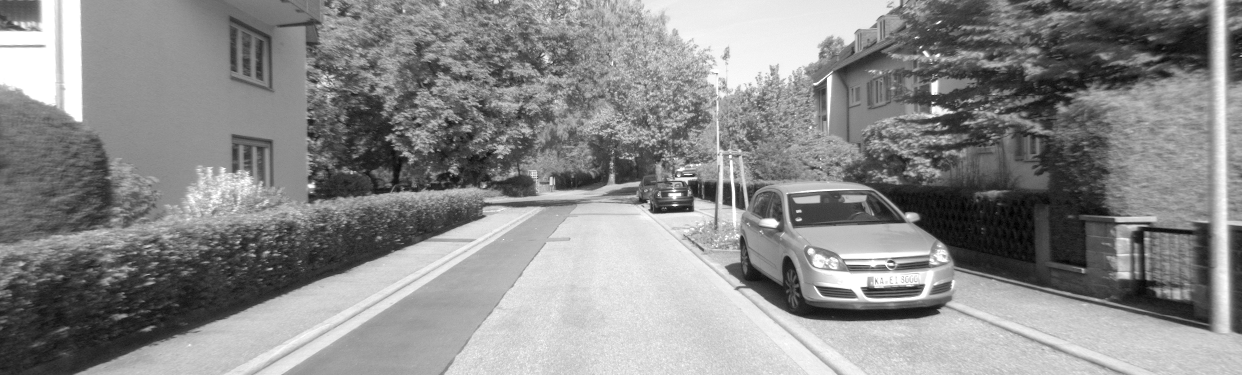}
\label{fig:gamma0.25}}
\qquad
\subfloat[][$\gamma > 1$.]{
\includegraphics[scale=0.12]{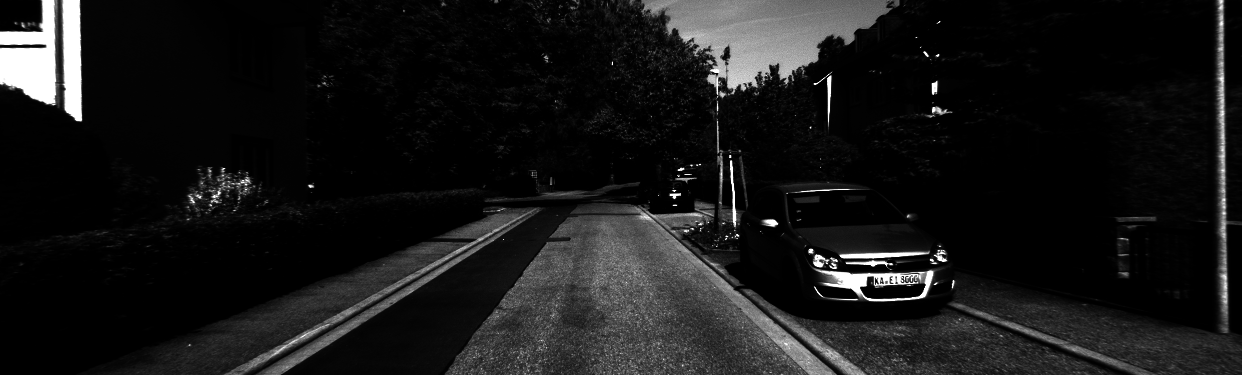}
\label{fig:gamma4}}
\caption{Examples of the distortions applied to KITTI to test the robustness of the algorithms.}
\label{fig:distortions}
\end{figure}

The gamma power transformation creates an image $I'$ from image $I$ by applying: $I' =I^{\gamma}$. As shown in Figure \ref{fig:distortions}, values of $\gamma < 1$ emulates camera overexposure and $\gamma > 1$ emulates camera underexposure. In our tests we used four values of gamma: $0.25$, $0.5$, $2$ and $4$. 

\section{RESULTS}

In this section, we present a quantitative and a qualitative comparison between the trajectories computed by each algorithm: LIFT-SLAM, LFNet-SLAM, and ORB-SLAM. The quantitative evaluation in KITTI sequences are based on the Relative Pose Error (RPE) of translation and rotation, as described in \cite{kitti-benchmark}, and Absolute Trajectory Error (ATE), detailed in \cite{tum-vi}. All results presented in this section were computed with a python package to evaluate odometry and SLAM called Evo \cite{evo}. Due to the size of the sequences, the estimates in Euroc were evaluated only by ATE. All of the quantitative metrics are an average of 5 executions due to the algorithms' stochastic nature. ORB-SLAM's results shown here were computed by our executions since, in ORB-SLAM's paper, an evaluation with RPE is not presented.

In table \ref{tab:all-results-kitti} we present the quantitative results for all algorithms in KITTI dataset. In sequences $01$ and $02$, none of the algorithms could compute the odometry for at least half of the sequence. Besides, LFNet-SLAM failed to track sequence $10$. However, it is possible to notice that the deep learning-based algorithms outperform ORB-SLAM in most sequences, especially in the smaller sequences, such as $03$ and $04$. Furthermore, LIFT-SLAM has achieved better performance than LFNet-SLAM in most of KITTI sequences. This is an indication that LIFT-SLAM has a better performance in outdoor environments than LFNet-SLAM.

\begin{table*}
\centering
\caption{Quantitative results in KITTI dataset. We fill with "X" the sequences unavailable due to tracking failure. The best average in each metric is highlighted and the second best is underlined.}
\label{tab:all-results-kitti}
\resizebox{0.9\textwidth}{!}{\begin{tabular}{|c|c|ccccccccccc|}
\hline \textbf{Algorithm} & \textbf{Metric} & \textbf{00} & \textbf{01} & \textbf{02} & \textbf{03} & \textbf{04} & \textbf{05} & \textbf{06} & \textbf{07} & \textbf{08} & \textbf{09} & \textbf{10}\\ \hline
               & ATE (m) & \underline{11.54}& X & X & 15.13& 4.29 & \textbf{7.74} & 20.26 & \underline{13.47} &\textbf{39.51} & \textbf{49.67} &\textbf{19.94}\\
ORB-SLAM      & $RPE_{trans}$ (\%) & \underline{4.46} & X & X & 9.75 & 3.71 & \textbf{3.35} & 8.11 & \underline{7.43} & \textbf{12.16} & 26.51 & \textbf{8.65} \\
              &$RPE_{rot}$ (deg/m) & 3.28 & X & X & 2.78 & 2.15 & \underline{3.57} & 2.88 & \textbf{3.58} &3.05 & 11.13 & \underline{3.62} \\ \hline

          & ATE (m)            &18.77 & X & X & \textbf{1.10} & \textbf{0.40} & \underline{8.09} & \underline{18.47} & \textbf{4.03} & \underline{80.97} & \underline{59.88} & \underline{31.84} \\
LIFT-SLAM & $RPE_{trans}$ (\%) & 6.71 & X & X & \textbf{0.87} & \textbf{2.10} & \underline{4.46} & \underline{7.76} &\textbf{2.51} & \underline{27.63} & \textbf{20.65} & \underline{10.08} \\
          & $RPE_{rot}$ (deg/m)& \textbf{2.20} & X & X & \textbf{0.34} & \textbf{0.65} & \textbf{2.58} & \underline{2.49} & \underline{3.60} & \underline{2.10} & \textbf{2.12} & \textbf{2.25} \\  \hline

          & ATE (m)           &\textbf{10.28} & X & X & \underline{2.57} & \underline{0.62} & 16.73 & \textbf{16.26} & 13.71 & 158.67 & 71.01 & X \\
LFNet-SLAM & $RPE_{trans}$ (\%) & \textbf{3.96} & X & X & \underline{1.56} & \underline{2.12} & 7.14 & \textbf{6.91} & 7.44 & 37.87 & \underline{26.03} & X \\
          & $RPE_{rot}$ (deg/m)& \underline{2.35} & X & X & \underline{0.51} & \underline{0.82} & 4.30 & \textbf{2.42} & 5.15 & \textbf{1.88} & \underline{3.20} & X \\  \hline
\end{tabular}}
\end{table*}

\begin{figure*}[t]
\centering
\subfloat[KITTI 00]{
\includegraphics[scale=0.108]{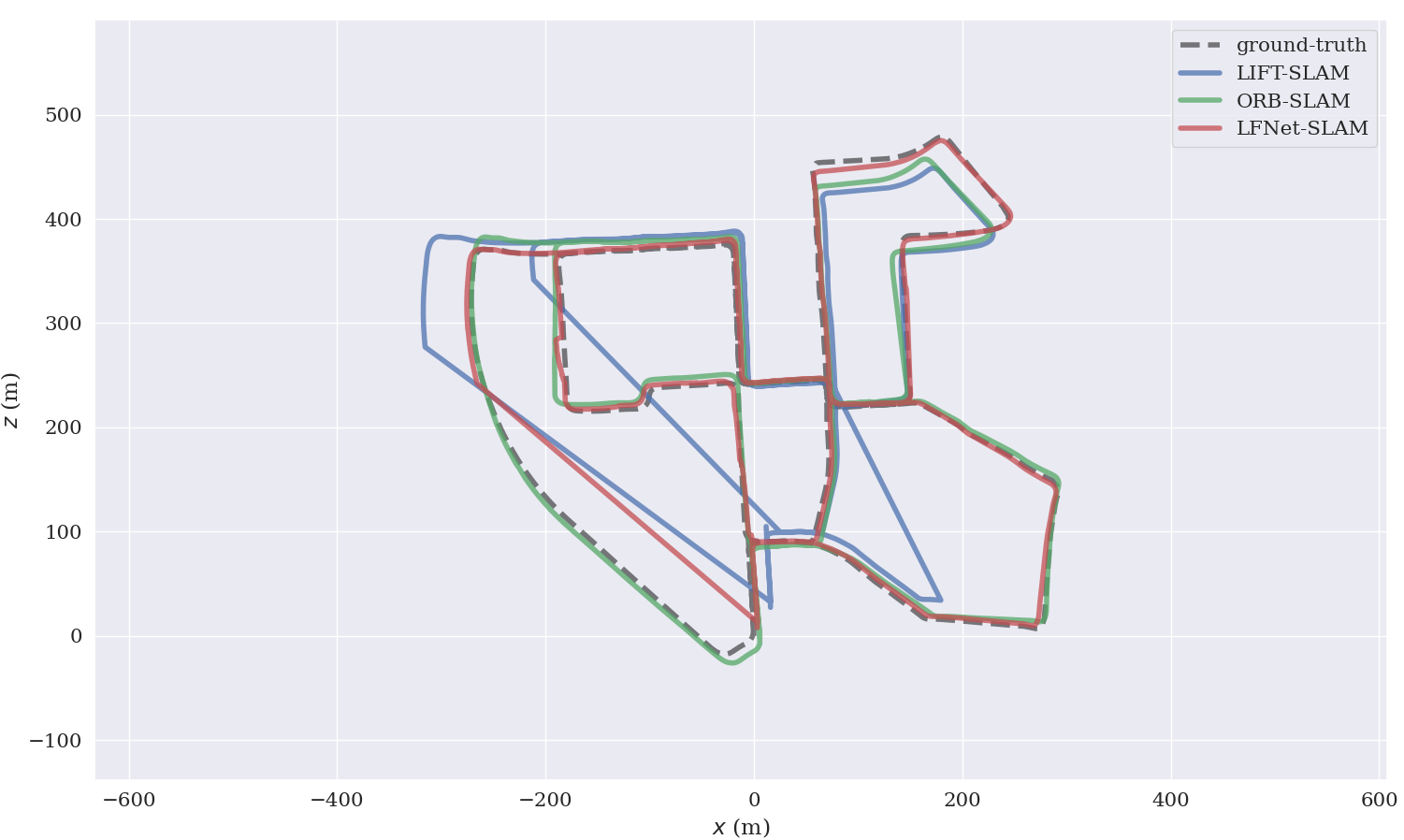}
\label{fig:kitti-00}}
\qquad
\subfloat[KITTI 03]{
\includegraphics[scale=0.15]{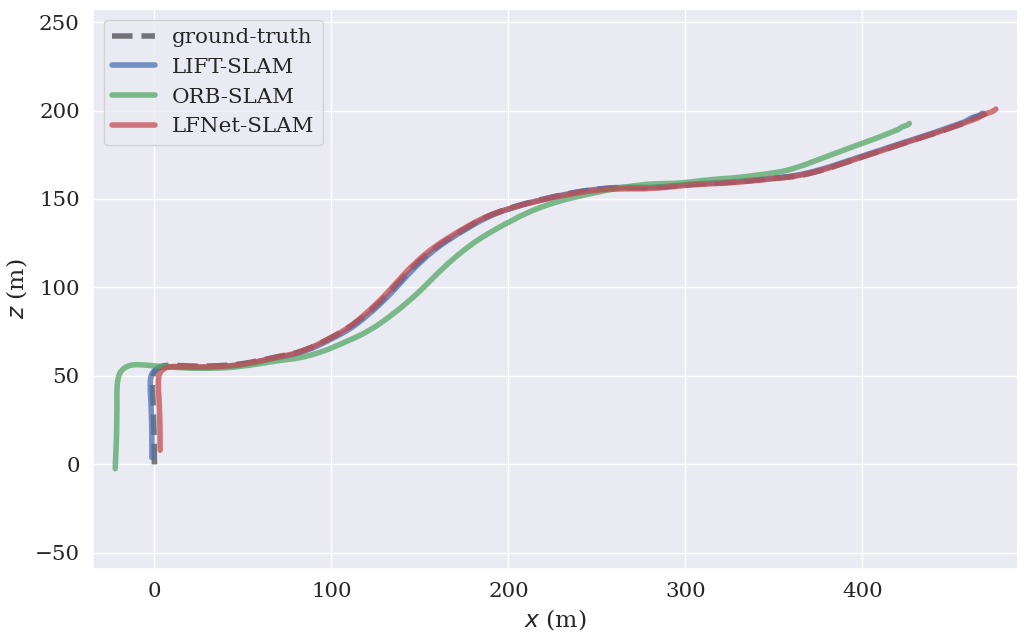}
\label{fig:kitti-03}}

\subfloat[KITTI 06]{
\includegraphics[scale=0.15]{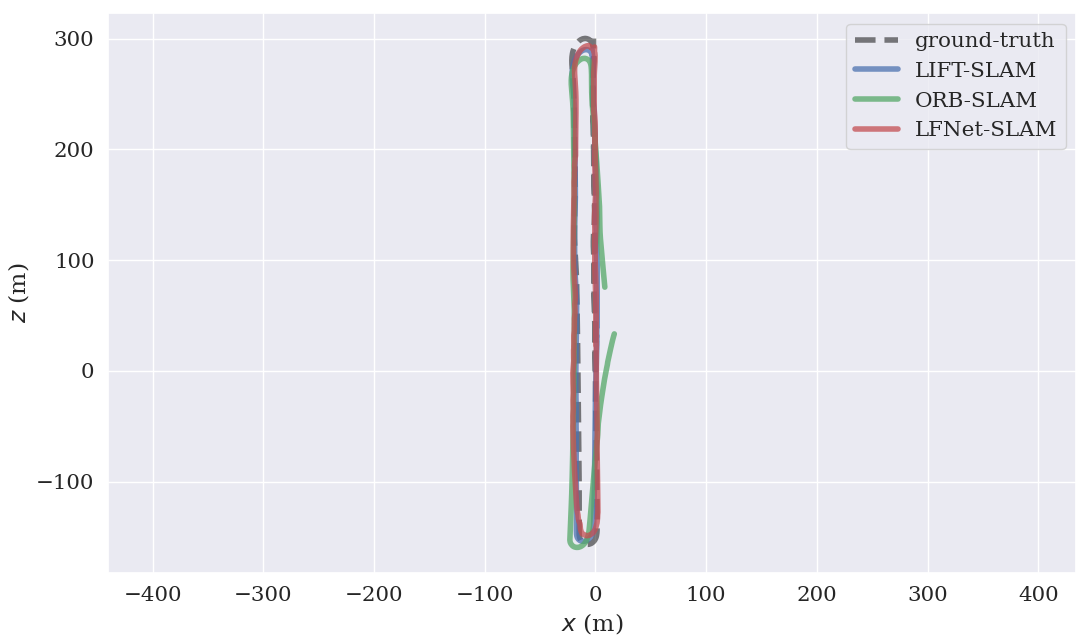}
\label{fig:kitti-06}}
\qquad
\subfloat[KITTI 07]{
\includegraphics[scale=0.15]{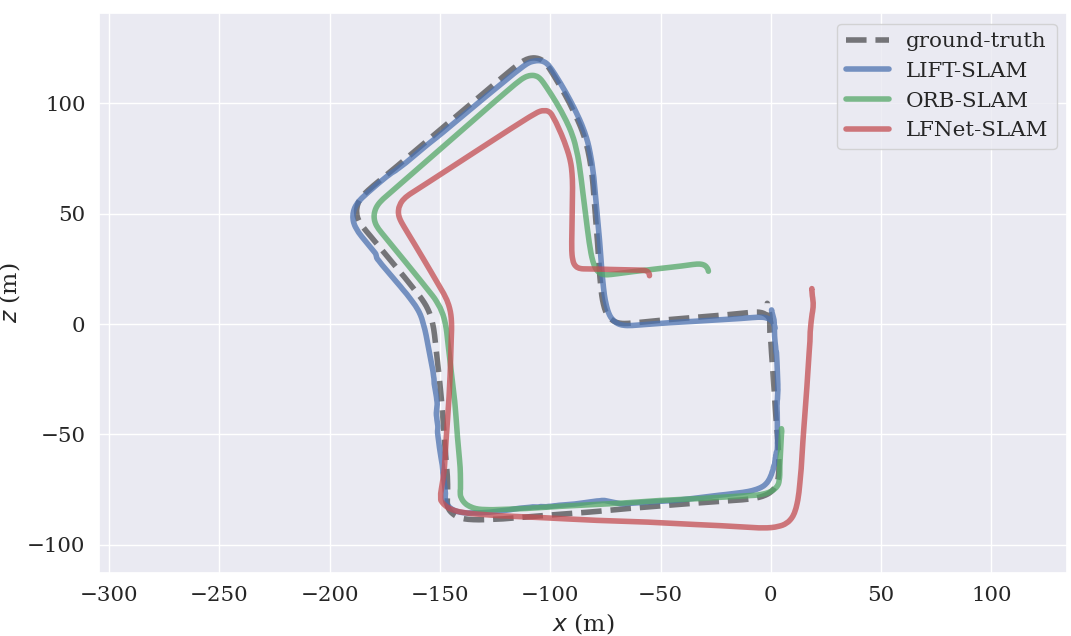}
\label{fig:kitti-07}}
\caption{Qualitative results in KITTI dataset.}
\label{fig:all-traj-kitti}
\end{figure*}

Figure \ref{fig:all-traj-kitti} shows the qualitative results of the algorithms in some KITTI sequences. In sequence $00$ (Figure \ref{fig:kitti-00}), we notice that ORB-SLAM was the only algorithm that did not lose track at some part of the trajectory. On the other hand, figures from \ref{fig:kitti-03} to \ref{fig:kitti-07} show that the ORB-SLAM's pose estimation suffers more from drift accumulation that the deep learning-based algorithms. LFNet-SLAM also has bad performance in sequence $07$ (Figure \ref{fig:kitti-07}), as it could not close the loop to correct the drift.

Table \ref{tab:all-results-euroc} presents the quantitative results of all algorithms in the Euroc MAV dataset. In this case, LIFT-SLAM could not track at least half of two sequences (V1\_01 and V1\_03). Moreover, LFNet-SLAM was the only algorithm that did not lose track in sequence V1\_03. Additionally, LFNet-SLAM performed better than LIFT-SLAM in all sequences. Therefore, LFNet-SLAM performs better in indoor environments than LIFT-SLAM.

\begin{table*}
\centering
\caption{Quantitative results in Euroc MAV dataset. We fill with "X" the sequences unavailable due to tracking failure. The best average in each metric is highlighted and the second best is underlined. ORB-SLAM results were obtained from \cite{orb-slam3}.}
\label{tab:all-results-euroc}
\resizebox{0.6\textwidth}{!}{\begin{tabular}{|c|cccccc|}
\hline \textbf{Algorithm} & \textbf{MH\_01} & \textbf{MH\_02} & \textbf{MH\_03} & \textbf{MH\_04} & \textbf{V1\_01} & \textbf{V1\_03} \\ \hline
ORB-SLAM & 0.071 & \underline{0.067} & \textbf{0.071} & \textbf{0.082} & \textbf{0.015} & X \\
LIFT-SLAM & \underline{0.062} & 0.227 & 0.144 & 1.859 & X & X \\
LFNet-SLAM & \textbf{0.041} & \textbf{0.045} & \underline{0.075} & \underline{1.852} & \underline{0.133} & \textbf{0.453} \\
\hline
\end{tabular}}
\end{table*}

\begin{figure*}[tb]
\centering
\subfloat[MH\_01]{
\includegraphics[scale=0.14]{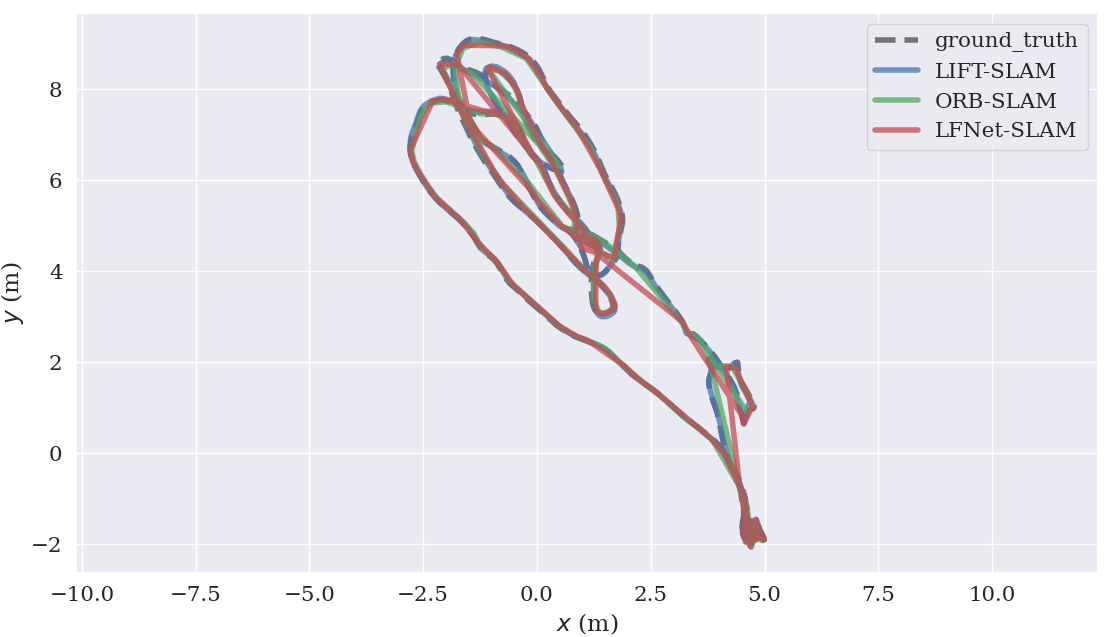}
\label{fig:mh01}}
\subfloat[MH\_02]{
\includegraphics[scale=0.14]{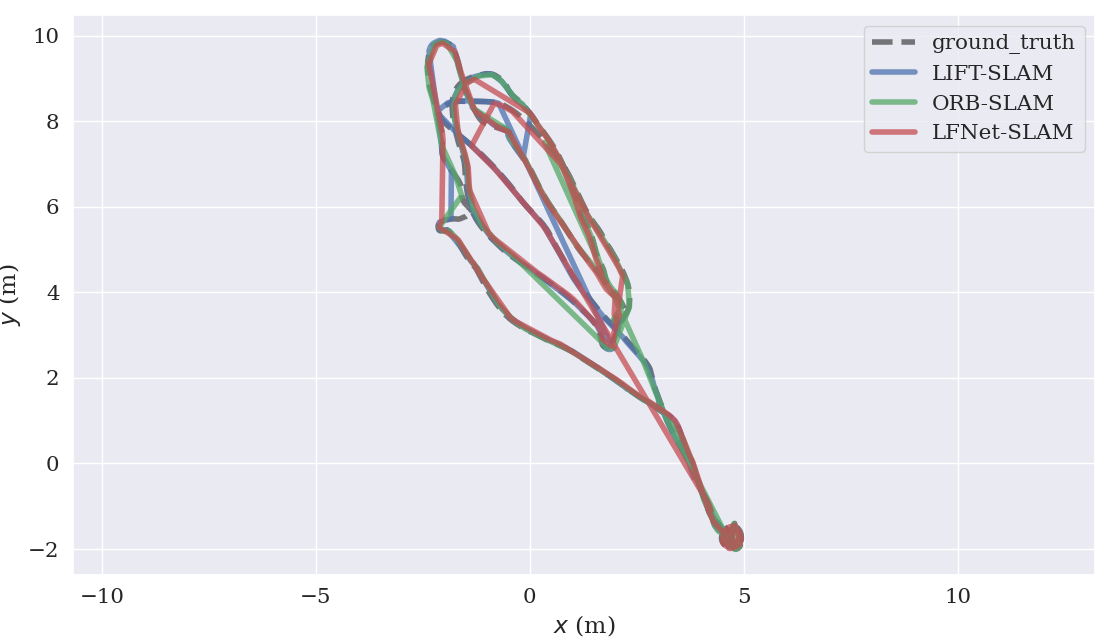}
\label{fig:mh02}}
\subfloat[MH\_04]{
\includegraphics[scale=0.14]{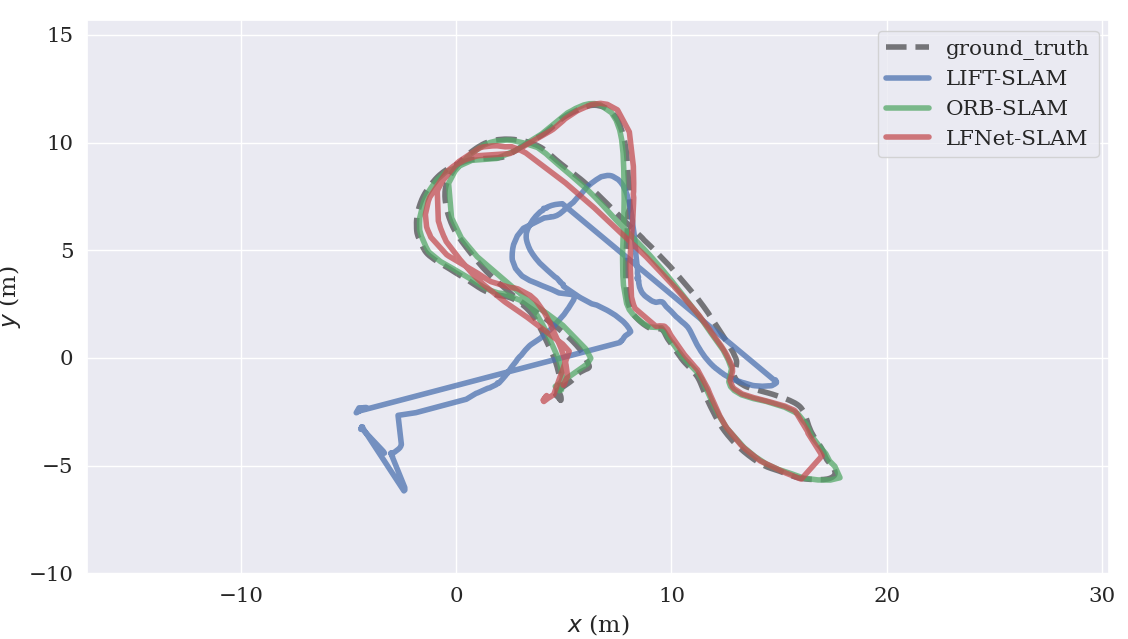}
\label{fig:mh04}}
\caption{Qualitative results in Euroc MAV dataset.}
\label{fig:all-traj-euroc}
\end{figure*}

Figure \ref{fig:all-traj-euroc} shows some qualitative results in Euroc MAV dataset. The performance of all the algorithms in sequence MH\_01 is similar. On the other hand, all algorithms failed to track some parts of sequence MH\_02. Furthermore, LIFT-SLAM accumulated a lot of drift in sequence MH\_04, as shown in Figure \ref{fig:mh04}.

Table \ref{tab:distortion-results} shows the results of the robustness tests (described in section \ref{sec:datasets}) in sequence KITTI $03$. We chose this sequence because it is one of the smaller sequences where all algorithms had a good performance and yet there are different types of motion and illumination changes, so we could analyze if the performance of the algorithms worsens under camera distortion.

In this case, we notice that ORB-SLAM could not perform VO under camera underexposing scenarios ($\gamma > 1$). None of the algorithms computed the pose for at least half of the sequence were $\gamma=4$. However, LIFT-SLAM and LFNet-SLAM successfully performed VO for $\gamma = 2$. We can also notice that in some cases, the algorithms even improved their performances with the distorted images. This occurs because the distortions remove some outliers from the images, and therefore, the algorithms can select better keypoints.

\begin{table*}[t]
\centering
\caption[]{Quantitative results of the robustness tests in KITTI $03$. We fill with "X" the sequences unavailable due to tracking failure.}
\label{tab:distortion-results}
\resizebox{\textwidth}{!}{\begin{tabular}{|c|ccc|ccc|ccc|}
\hline
 \textbf{Distortion} &  \multicolumn{3}{|c|}{\textbf{ORB-SLAM}} &  \multicolumn{3}{c|}{\textbf{LIFT-SLAM}} & \multicolumn{3}{c|}{\textbf{LFNet-SLAM}} \\ \hline
 &  $\mathbf{RPE_{trans}}$ \textbf{(\%)} & $\mathbf{RPE_{rot}}$ \textbf{(deg/m)} & \textbf{ATE (m)} & $\mathbf{RPE_{trans}}$ \textbf{(\%)} & $\mathbf{RPE_{rot}}$ \textbf{(deg/m)} & \textbf{ATE (m)} & $\mathbf{RPE_{trans}}$ \textbf{(\%)} & $\mathbf{RPE_{rot}}$ \textbf{(deg/m)} & \textbf{ATE (m)} \\ \hline
  no distortion & 9.75 & 2.78 &15.13 & 0.87 & 0.34 & 1.10 & 1.56 & 0.51 & 2.57\\
  $\gamma = 0.25$ & 7.68 & 1.95 & 11.72 & 0.98 & 0.43 & 1.16 & 6.50 & 13.98 & 10.50 \\
 $\gamma = 0.5$ & 8.25 & 2.24 & 11.38 & 0.91 & 0.36 & 1.06 & 0.88 & 0.43 & 1.30\\
  $\gamma = 2$ & X & X & X & 1.62 & 0.62 & 2.38 & 1.28 & 0.55 & 2.00 \\
 $\gamma = 4$ & X & X & X & X & X & X & X & X & X\\
\hline
\end{tabular}}
\end{table*}

\begin{figure*}[tb]
\centering
\subfloat[ORB-SLAM]{
\includegraphics[scale=0.14]{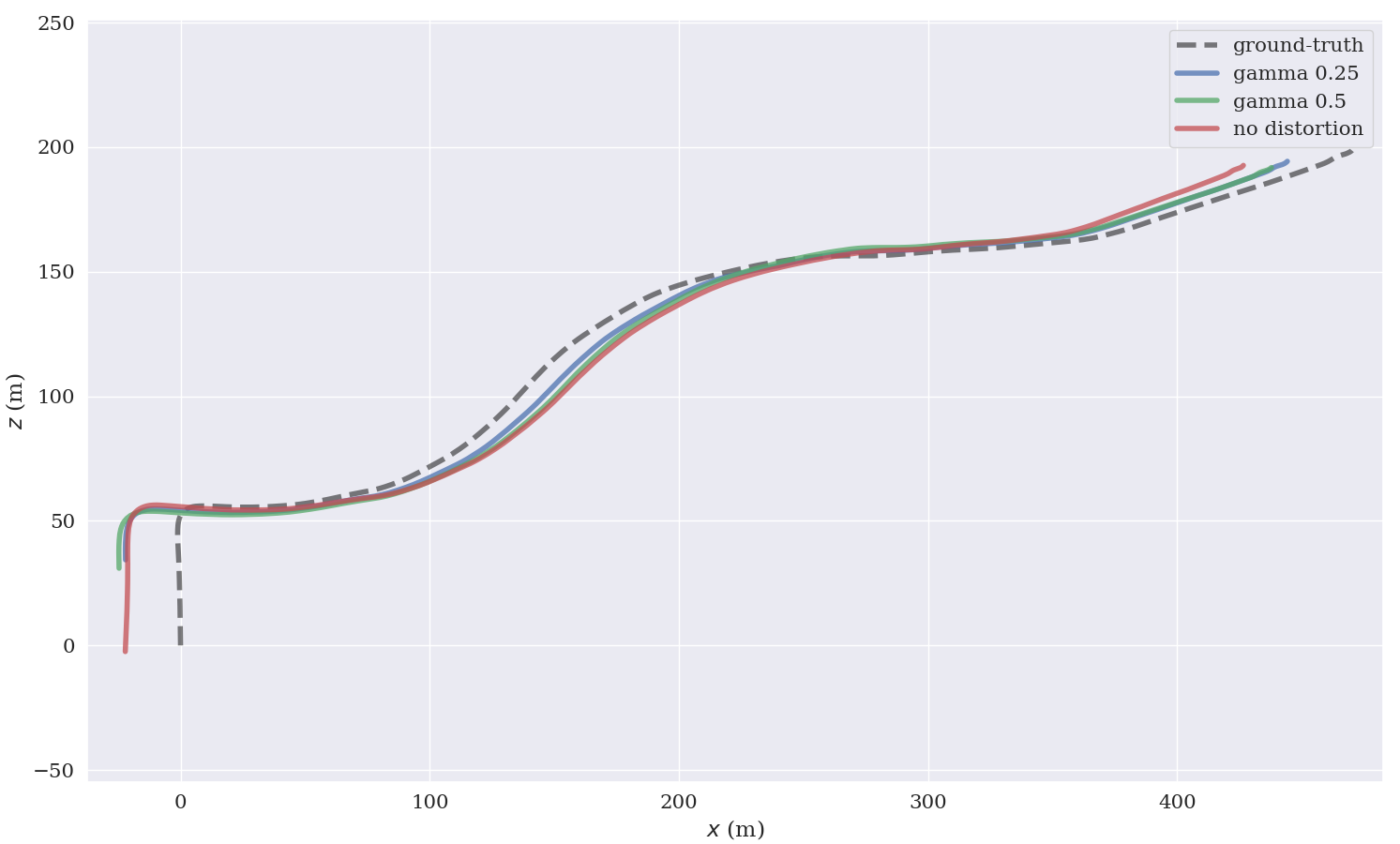}
\label{fig:noise-orb}}
\subfloat[LIFT-SLAM]{
\includegraphics[scale=0.14]{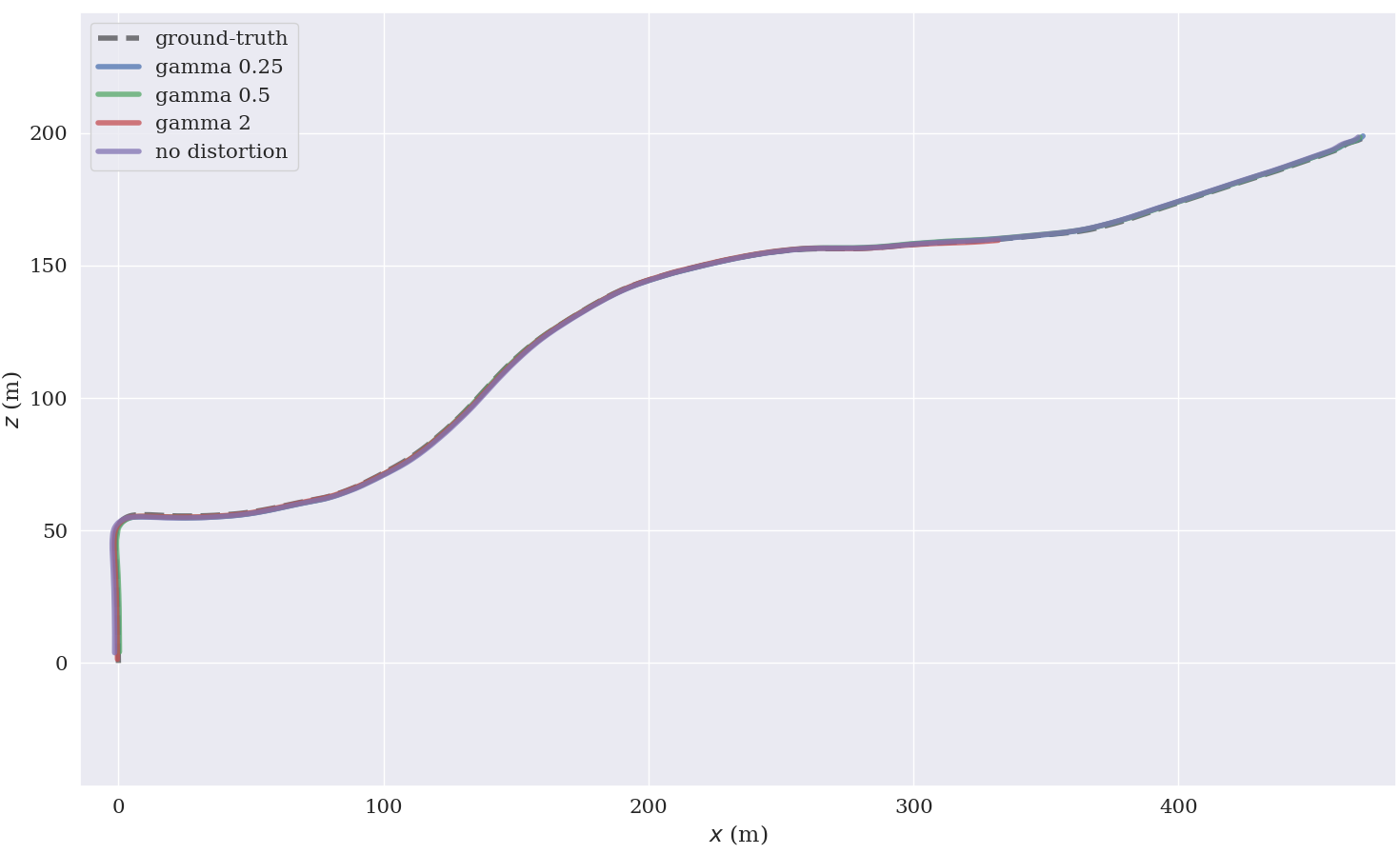}
\label{fig:noise-lift}}
\qquad
\subfloat[LFNet-SLAM]{
\includegraphics[scale=0.14]{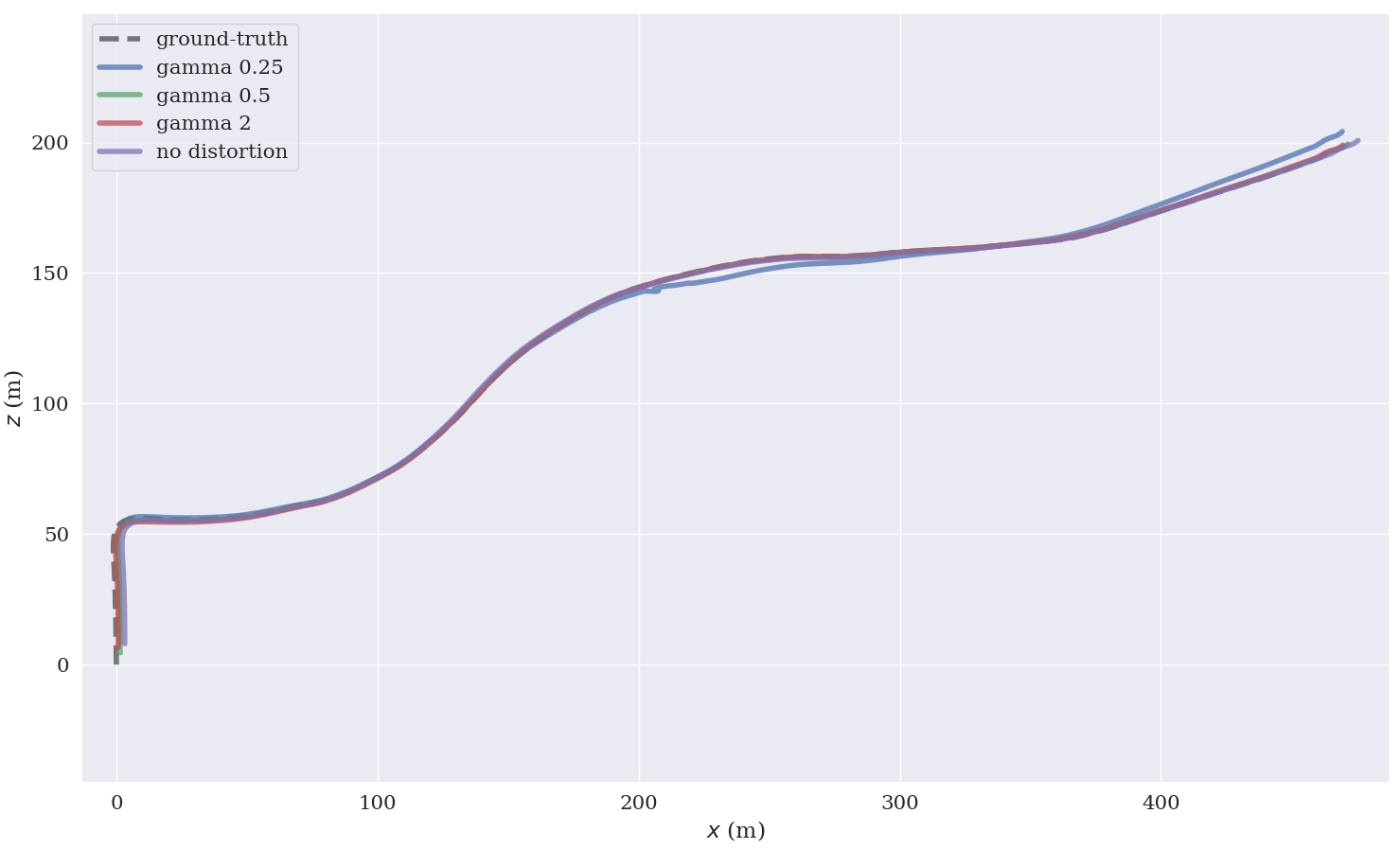}
\label{fig:noise-lfnet}}
\caption{Qualitative results of the robustness tests.}
\label{fig:noise-trajs}
\end{figure*}

Figure \ref{fig:noise-trajs} shows a qualitative comparison between the trajectories computed without distortion and under distortion for each algorithm. The trajectories computed by LIFT-SLAM under distortion are similar to the trajectory without distortion. Therefore, the algorithm was robust to the camera noise in this scenario. On the other side, in LFNet-SLAM, the trajectory is worse for $\gamma = 0.25$.

Although we noticed that the deep learning-based VSLAM algorithms could be more robust than a traditional VSLAM algorithm, they are still not capable of estimating the pose in some cases. In \cite{lift-slam}, we show that fine-tuning the DNNs with VO sequences can improve the robustness and the accuracy of the deep learning-based algorithms. This is possible because, with fine-tuning, the DNNs can learn characteristics presented only in VO sequences.

\section{CONCLUSIONS}

In this paper, we presented an evaluation of deep learning-based hybrid VSLAM algorithms. We also showed a comparison between these algorithms and ORB-SLAM. Based on RPE and APE, we concluded that the algorithms with learned features improved ORB-SLAM accuracy in most of KITTI and Euroc Sequences. Moreover, LIFT-SLAM presented a better performance of the odometry in the outdoor environments than LFNet-SLAM. On the other side, LFNet-SLAM computed the pose more accurately in indoor environments. These results are influenced by the datasets used to train each DNN.

Furthermore, the deep learning-based algorithms were more robust to the camera ill exposure conditions we emulated in sequence KITTI $03$. In this case, LIFT-SLAM was less affected by the camera noise than LFNet-SLAM. In some scenarios, the algorithms even improved their results with the distortions applied to the images. This fact confirms that a good feature selection can improve the performance of the VSLAM algorithms. Therefore, in future work, we plan to develop an attention-based mechanism to select features for VSLAM.

Although the learned features improved the performance of a traditional VSLAM algorithm, fine-tuning the DNNs with VO sequences can improve the robustness and the accuracy of the deep learning-based algorithms. Therefore, in future work, we consider to fine-tune both LIFT and LFNet with VO sequences.





\section*{ACKNOWLEDGMENT}
This work was supported by the Brazilian National Council for Scientific and Technological Development (CNPq), Coordination for the Improvement of Higher Education Personnel (CAPES) and the company Quinto Andar.


\bibliographystyle{IEEEtran}
\bibliography{IEEEabrv,references}

\end{document}